\documentclass[11pt,a4paper]{article}
\usepackage[hyperref]{emnlp2020}
\usepackage{times}
\usepackage{latexsym}

\usepackage{microtype}

\aclfinalcopy 
\def\aclpaperid{190} 

\usepackage{graphicx}
\usepackage{booktabs}
\usepackage{multirow}
\usepackage{amsmath}
\usepackage{amssymb}
\usepackage{bm}
\usepackage{url}
\usepackage{floatrow}
\usepackage{xcolor}
\newfloatcommand{capbtabbox}{table}[][\FBwidth]

\title{Difference-aware Knowledge Selection for Knowledge-grounded Conversation Generation}

\author{Chujie Zheng$^\dagger$, Yunbo Cao$^\ddagger$, Daxin Jiang$^\S$, Minlie Huang$^\dagger$\thanks{* Corresponding author.} \\
  $^\dagger$Department of Computer Science and Technology, $^\dagger$Institute for Artificial Intelligence, \\
  $^\dagger$State Key Lab of Intelligent Technology and Systems, \\
  $^\dagger$Beijing National Research Center for Information Science and Technology, \\
  $^\dagger$Tsinghua University, Beijing 100084, China \\
  $^\ddagger$Smart Platform Product Department, Tencent, Beijing \\
  $^\S$STCA NLP Group, Microsoft, Beijing \\
  {\tt chujiezhengchn@gmail.com, yunbocao@tencent.com} \\
  {\tt djiang@microsoft.com, aihuang@tsinghua.edu.cn} \\
}

\date{}

\begin{document}
\maketitle
\begin{abstract}

  In a multi-turn knowledge-grounded dialog, the difference between the knowledge selected at different turns usually provides potential clues to knowledge selection, which has been largely neglected in previous research.
  In this paper, we propose a difference-aware knowledge selection method. It first computes the difference between the candidate knowledge sentences provided at the current turn and those chosen in the previous turns.
  Then, the differential information is fused with or disentangled from the contextual information to facilitate final knowledge selection.
  Automatic, human observational, and interactive evaluation shows that our method is able to select knowledge more accurately and generate more informative responses, significantly outperforming the state-of-the-art baselines.
  {The codes are available at \url{https://github.com/chujiezheng/DiffKS}.}

\end{abstract}

\section{Introduction}
\label{sec:introduction}


Knowledge-grounded conversation generation aims at generating informative responses based on both discourse context and external knowledge \cite{ghazvininejad2018knowledge,zhou-2018-commensense}, 
where selecting appropriate knowledge is critical to the success of the task.
Existing knowledge selection models generally fall into two types.
One type is solely based on the context \cite{postks-ijcai,zhang-2019-improving,meng-2020-refnet,ren-2020-glks}, which we call \textbf{non-sequential selection} because knowledge selection at different turns is independent.
The other type sequentially selects knowledge additionally conditioned on previously selected knowledge \cite{Kim2020Sequential}, which we call \textbf{sequential selection}. 
As shown in \citet{Kim2020Sequential}, such a sequential way can better simulate a multi-turn dialog and facilitate knowledge selection in later turns.

\begin{figure}[t]
  \centering
  \includegraphics[width=0.9\linewidth]{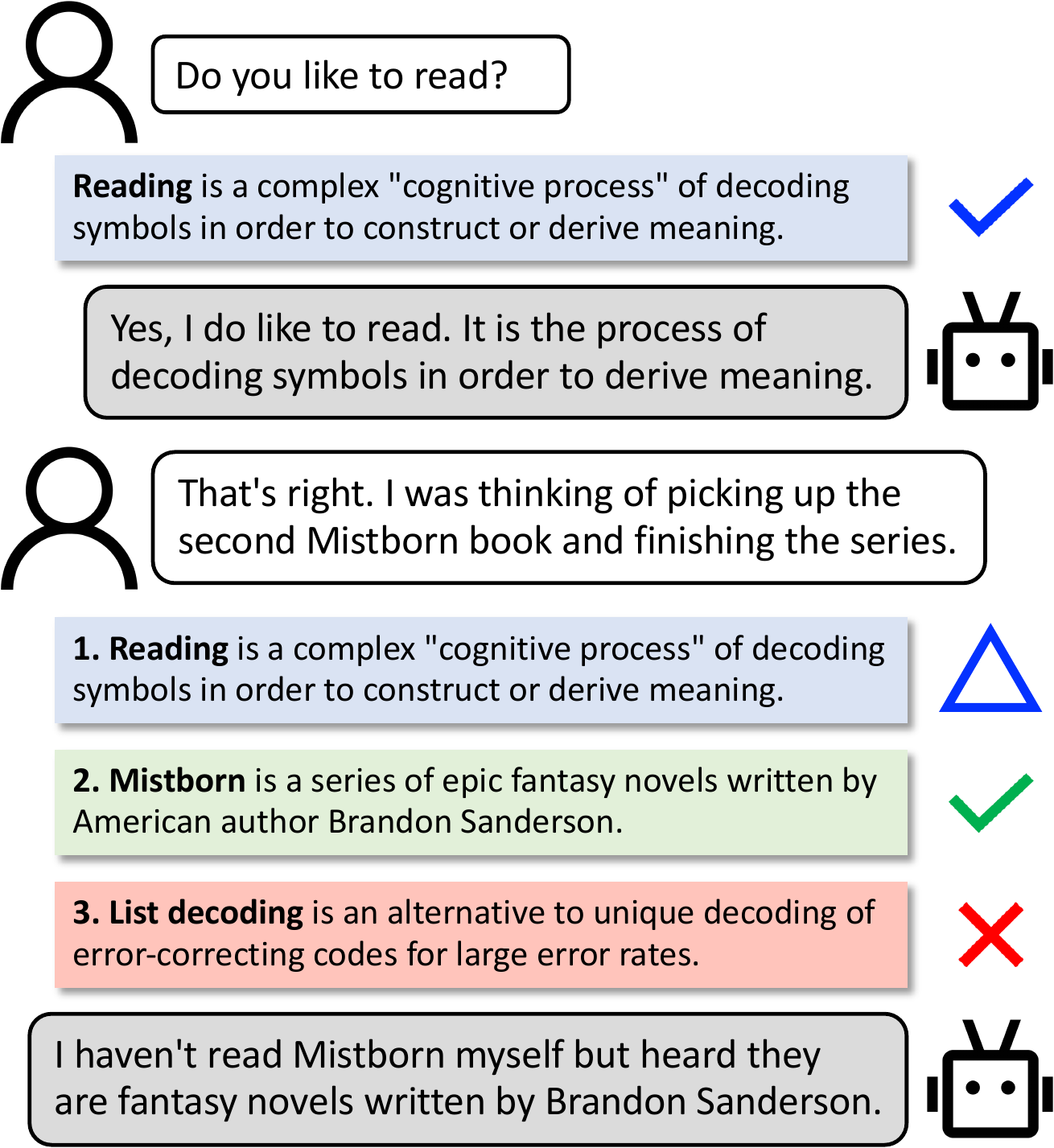}
  \caption{An example of difference-aware knowledge selection.
  The {\color{blue} \textbf{blue} $\bm{\triangle}$} denotes that the corresponding knowledge has little difference from or is identical to the previously selected one, and selecting it may lead to \textit{repetitive} responses. 
  The {\color{red} \textbf{red} $\bm{\times}$} denotes that the difference is too large, and selecting it could make the response \textit{incoherent} with the context.
  }
  \label{fig:intro_example}
\end{figure}

However, the \textbf{difference} between selected knowledge at different turns has been largely neglected in prior studies, while it usually provides potential clues to knowledge selection.
Figure \ref{fig:intro_example} illustrates an example, where the dialog system selects one from candidate knowledge sentences (all relevant to the context) at the 2$^\text{nd}$ turn.
Selecting the knowledge that has little difference from or even is identical to the previously selected one (like the 1$^\text{st}$ knowledge)
may lead to generating \textit{repetitive} responses, 
while too large difference (like the 3$^\text{rd}$ knowledge) would make the response \textit{incoherent} with the context.
As a result, the dialog system should avoid the knowledge which differs from the previously selected ones either too little or too largely, and instead select an appropriate knowledge sentence (the 2$^\text{nd}$ one) which can make the conversation flow smoothly and naturally.


We thus propose DiffKS, a novel \textbf{Diff}erence-aware \textbf{K}nowledge \textbf{S}election method for knowledge-grounded conversation generation.
It first computes the difference between the candidate knowledge sentences provided at the current turn and the previously selected knowledge.
Then, in the two models we devise, the differential information is fused with or disentangled from the contextual information to facilitate final knowledge selection.
Automatic and human evaluation on two widely-used benchmarks shows that our method is significantly superior over the state-of-the-art baselines and it can select knowledge more accurately and generate more informative responses.

Our contributions are summarized as follows:
\begin{itemize}

    \item We propose to explicitly model and utilize the differential information between selected knowledge in multi-turn knowledge-grounded conversation for knowledge selection.
    We further devise two variants where the differential information is fused with or disentangled from the context information during knowledge selection.
    
    \item Automatic, human observational, and human interactive evaluations show that our method significantly outperforms strong baselines in terms of knowledge selection and can generate more informative responses.

\end{itemize}

\section{Related Work}
\label{sec:related_work}

\subsection{Knowledge-grounded Dialog Generation}

Recently, a variety of neural models have been proposed to facilitate knowledge-grounded conversation generation \cite{zhu2017flexible,young-2018-aaai,zhou-2018-commensense,liu-etal-2018-knowledge}.
The research topic is also greatly advanced by many corpora \cite{zhou-etal-2018-dataset,moghe-etal-2018-towards,dinan2018wizard,Gopalakrishnan2019,moon-etal-2019-opendialkg,tuan-etal-2019-dykgchat,wu-etal-2019-proactive,zhou-etal-2020-kdconv}. 
As surveyed in \citet{huang-etal-2019-challenges}, existing studies have been mainly devoted to addressing two research problems:
(1) \textit{\textbf{knowledge selection}}: selecting appropriate knowledge given the dialog context and previously selected knowledge \cite{postks-ijcai,zhang-2019-improving,meng-2020-refnet,ren-2020-glks,Kim2020Sequential};
and (2) \textit{\textbf{knowledge-aware generation}}: injecting the required knowledge to generate meaningful and informative responses \cite{ghazvininejad2018knowledge,zhou-2018-commensense,li-etal-2019-incremental,qin-etal-2019-conversing,yavuz-etal-2019-deepcopy,Zhao2020Low-Resource}. 
Since selecting the appropriate knowledge is a precursor to the success of knowledge grounded dialog systems, we focus on the \textit{\textbf{knowledge selection}} problem in this paper.

\subsection{Non-sequential Knowledge Selection}

The non-sequential selection models capture the relationship between the current context and background knowledge \cite{postks-ijcai,zhang-2019-improving,meng-2020-refnet,ren-2020-glks}.
For instance, PostKS \cite{postks-ijcai} estimates a posterior distribution over candidate knowledge sentences, which is based on both the context and the golden response, and only uses the context to estimate a prior distribution as an approximation of the posterior during inference.

Besides, \citet{zhang-2019-improving,meng-2020-refnet,ren-2020-glks} also belong to non-sequential selection models.
Different from our work and \citet{postks-ijcai,Kim2020Sequential} that select knowledge from \textit{candidate knowledge sentences}, their methods are devised for selecting important text spans or fragments from the \textit{background knowledge document} that will be used in generation.
Therefore these works have a different task setting from ours.

\subsection{Sequential Knowledge Selection}

The sequential selection models additionally make use of previously selected knowledge to facilitate knowledge selection \cite{Kim2020Sequential}.
For instance, \citet{Kim2020Sequential} propose a Sequential Latent Knowledge Selection (SLKS) model. 
It keeps track of the hidden states of dialog history and previously selected knowledge sentences.
Our method is parallel to SLKS because we also utilize the previously selected knowledge.
However, we explicitly compute the difference between knowledge selected at different turns, while SLKS only encodes the already selected knowledge in an implicit way.

In addition, recently there emerge a number of works that propose RL-based models to select a path in \textit{structured knowledge graph (KG)} \cite{xu-2020-aaai,xu-2020-acl}, which also select knowledge in a sequential way.
While our method is designed to ground the conversation to \textit{unstructured knowledge text},
we will leave as future work the application of our method to such KG-grounded dialog generation tasks \cite{wu-etal-2019-proactive,moon-etal-2019-opendialkg,zhou-etal-2020-kdconv}.


\section{Methodology}

\subsection{Task Formulation}

In a multi-turn dialogue, given a post and a sequence of knowledge sentences at each turn, our goal is to select appropriate knowledge and generate a proper response to the current context. 

Formally, the post at the $\tau$-th turn is a sequence of tokens $\mathrm{x}^{\tau} = \left( \mathrm{x}^{\tau}_1, \dots, \mathrm{x}^{\tau}_{\left| \mathrm{x}^{\tau} \right|} \right)$, and the response to be generated is $\mathrm{y}^{\tau} = \left( \mathrm{y}^{\tau}_1, \dots, \mathrm{y}^{\tau}_{\left| \mathrm{y}^{\tau} \right|} \right)$.
The background knowledge $\mathrm{k}^{\tau} = \left( \mathrm{k}^{\tau}_1, \dots, \mathrm{k}^{\tau}_{\left| \mathrm{k}^{\tau} \right|} \right) $ contains a sequence of knowledge sentences provided at the $\tau$-th turn. 
For each $i$, $\mathrm{k}^{\tau}_{i} = \left( \mathrm{k}^{\tau}_{i,1}, \dots, \mathrm{k}^{\tau}_{i,\left| \mathrm{k}_{i}^{\tau} \right|} \right) $ is a sequence of tokens in the $i$-th sentence.

Note that under the setting of multi-turn dialogue, we use $\mathrm{c}^{\tau} \triangleq \left[ \mathrm{x}^{\tau-1}; \mathrm{y}^{\tau-1}; \mathrm{x}^{\tau} \right]$ as the given context at the $\tau$-th turn, where $[\cdot; \cdot]$ denotes concatenation.
In Section \ref{sec:encoder} and \ref{sec:decoder}, we will omit the superscript $\tau$ for simplicity.

\subsection{Encoders}
\label{sec:encoder}

The context is encoded with a bidirectional GRU \cite{cho-etal-2014-learning}:
\begin{align}
  \left({\bm{h}}_{c,1}, \dots, {\bm{h}}_{c,\left| \mathrm{c} \right|} \right) = {\bm{BiGRU}}_c \left( \mathrm{c} \right),
\end{align}
where ${\bm{h}}_{c,i} = \left[ \overrightarrow{\bm{h}}_{c,i}; \overleftarrow{\bm{h}}_{c,i} \right]$. 
We use $\bm{h}_{c} \triangleq \left[ \overrightarrow{\bm{h}}_{c,\left| \mathrm{c} \right|}  ; \overleftarrow{\bm{h}}_{c,1} \right]$ as the context representation. 
Similarly, the knowledge sentences are encoded with another BiGRU:
\begin{align}
  \left(\bm{h}_{k,i,1}, \dots, \bm{h}_{k,i,\left|\mathrm{k}_{i}\right|} \right) = {\bm{BiGRU}}_{k} \left( \mathrm{k}_{i} \right). 
\end{align}
We use $\bm{h}_{k,i} \triangleq \left[ \overrightarrow{\bm{h}}_{k,i,\left|\mathrm{k}_{i}\right|} ; \overleftarrow{\bm{h}}_{k,i,1} \right]$ as the representation of $\mathrm{k}_{i}$. 
Specifically, we add an empty sentence $\mathrm{k}_0$ that indicates \textit{no knowledge being used}.

\subsection{Difference-aware Knowledge Selection}

In order to select proper knowledge, our model gets aware of the difference between the current candidate knowledge sentences and the previously selected knowledge. 

To make full use of the contextual dependency and relevance between the knowledge sentences\footnote{
  For example, the knowledge sentences may be extracted from a document in order, or about the same topic like in Wizard of Wikipedia \cite{dinan2018wizard}.
}, our model first compares candidate knowledge sentences to explore their correlations, where the comparison is conducted using BiGRU:
\begin{align}
  \left({\bm{r}}^{\tau}_{0}, \dots, {\bm{r}}^{\tau}_{ \left| {\mathrm{k}}^{\tau} \right| } \right)\! =\! \bm{BiGRU}\! \left( { \bm{h}}^{\tau}_{k,0}, \dots, { \bm{h}}^{\tau}_{k, \left| {\mathrm{k}}^{\tau} \right| } \right)\!,
\end{align}

\begin{figure}[t]
  \centering
  \includegraphics[width=\linewidth]{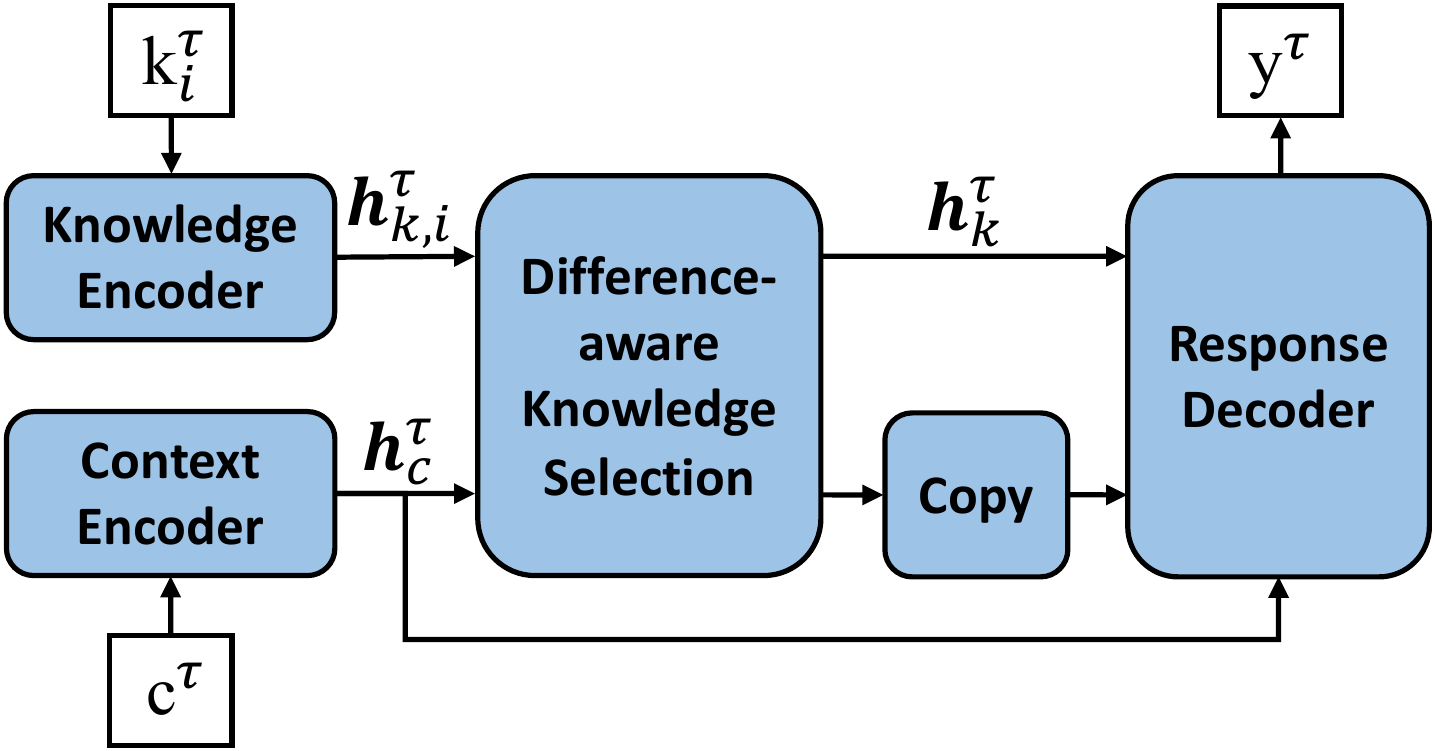}
  \caption{An overview of model structure.
  }
  \label{model_figure}
\end{figure}

Then, the model computes the difference of each knowledge sentence $\bm{r}^{\tau}_i$ from 
the knowledge selected in the previous $M$ turns  $\left\{ \bm{h}_{k}^{\tau-m} \right\}_{m=1}^{M}$:
\begin{align}
  \bm{o}^{\tau}_i &= \sum_{m=1}^M 
  \lambda_m \bm{Diff} \left( \bm{h}_{k}^{\tau-m}, \bm{r}^{\tau}_i \right), \label{equ:diff_M} \\
  & \sum_{m=1}^M \lambda_m = 1,\ \ \forall m,\lambda_m \ge 0 
\end{align}
Inspired by \citet{wang-etal-2018-co}, we define the difference as follow:
\begin{align}
  \bm{Diff} \left( \bm{x}, \bm{y} \right) \triangleq \bm{F}\left( \left[ \bm{x} - \bm{y} ; \bm{x} \odot \bm{y} \right] \right),
\end{align}
where $\bm{F}$ is a fully connected layer activated with $\mathrm{tanh}$.
Note that at the first turn, we set $\bm{o}^{1}_i$ to a zero vector because there is no differential information to be obtained.

For that intuitively the knowledge selected in the previous turn has the largest impact and most clues for the current selection, we studied the simplest case where $M=1$, saying $\bm{o}^{\tau}_i = \bm{Diff} \left( \bm{h}_{k}^{\tau-1}, \bm{r}^{\tau}_i \right)$, in the main experiments for simplicity.

Next, we introduce two variants where the differential information $\left\{\bm{o}^{\tau}_i\right\}_{i=0}^{\left| {\mathrm{k}}^{\tau} \right|}$ is fused with or disentangled from the contextual information during knowledge selection.

\subsubsection{Fused Selection}
\label{sec:fused_selection}

\begin{figure}[ht]
  \centering
  \includegraphics[width=0.7\linewidth]{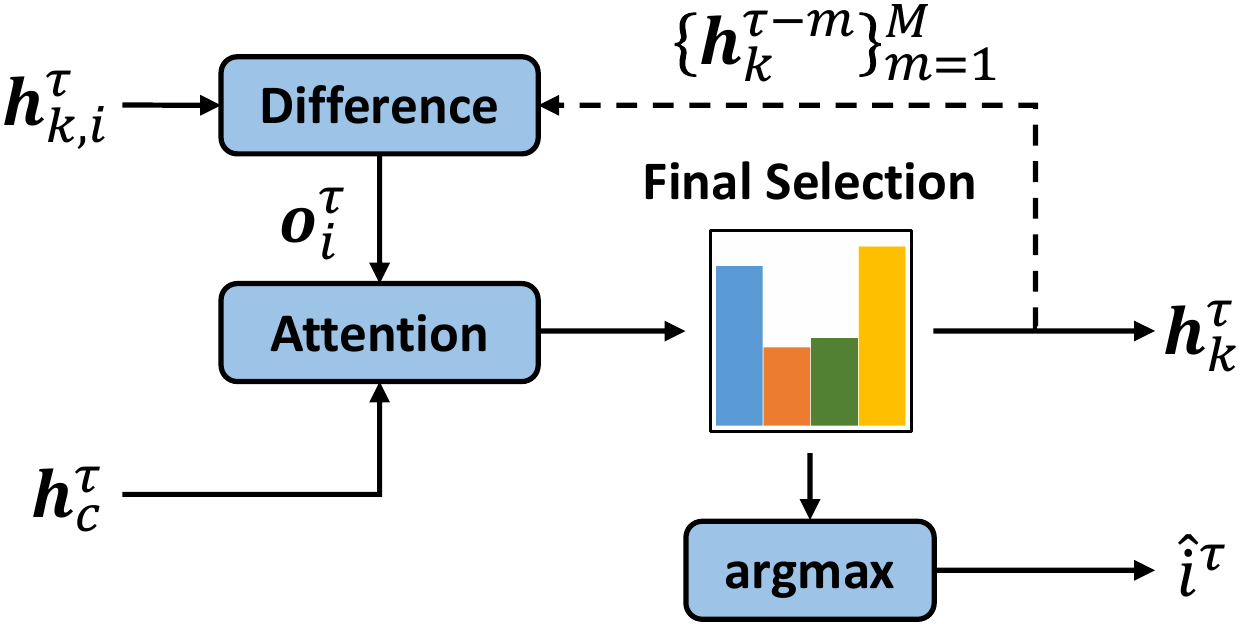}
  \caption{Fused Selection module.
  The contextual information and the differential information are \textit{fused} together to calculate the final knowledge selection distribution.}
  \label{fig:fused_selection}
\end{figure}

The Fused Selection module is shown in Figure \ref{fig:fused_selection}.
Directly taking $\bm{o}^{\tau}_i$ as an extra feature of $\mathrm{k}^{\tau}_i$, it uses the context $\bm{h}_{c}^{\tau}$ to query the difference-enhanced knowledge sentences:
\begin{align}
  \beta^{\tau}_{i}\! =\! \bm{v}^\mathrm{T}\tanh\left(\bm{W}_{\mathrm{que}} \bm{h}_{c}^{\tau}\! +\! \bm{W}_{\mathrm{key}} \left[ \bm{h}_{k,i}^{\tau} ; \bm{o}^{\tau}_i \right] \right)\!,
\end{align}
where $\bm{v}$, $\bm{W}_{\mathrm{que}}$ and $\bm{W}_{\mathrm{key}}$ are trainable parameters. 

However, it is difficult to distinguish the respective contributions of contextual and differential information to knowledge selection in the above fused variant.
We thus devise the disentangled variant as following, where the roles of two types of information are separated, which makes it feasible to conduct ablation study.

\subsubsection{Disentangled Selection}

\begin{figure}[ht]
  \centering
  \includegraphics[width=\linewidth]{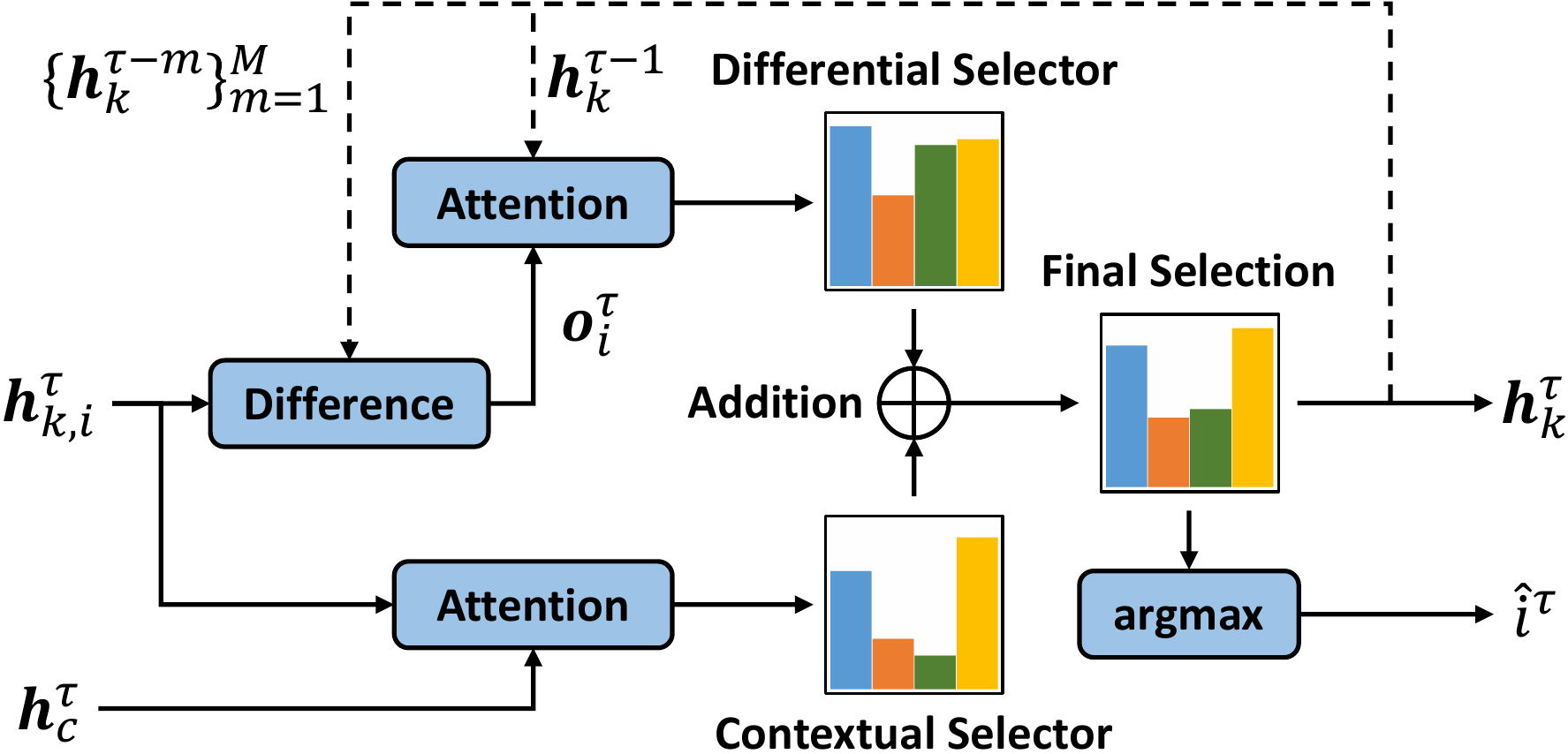}
  \caption{Disentangled Selection module.
  The contextual information and the differential information are \textit{disentangled} to calculate two separate knowledge selection distributions in two independent selectors.}
  \label{fig:disentangled_selection}
\end{figure}

Figure \ref{fig:disentangled_selection} gives an overview of the Disentangled Selection module.
It has two independent selectors. 
The \textbf{Contextual Selector} simply looks for the knowledge sentence that has high relevance to the context, just like most existing knowledge selection models do.
It only takes advantage of the context $\bm{h}_{c}^{\tau}$ to match each knowledge sentence itself $\bm{h}_{k,i}^{\tau}$, obtaining a context-aware selection distribution:
\begin{align}
  \beta^{\tau}_{\mathrm{Ctx},i} = \left( \bm{h}_{c}^{\tau} \right)^{\mathrm{T}} \bm{h}_{k,i}^{\tau}.
\end{align}

In contrast, the \textbf{Differential Selector} focuses on predicting the next knowledge to be selected conditioned on the previously selected knowledge and differential information, which reveals the process of knowledge transition.
Without the access to the contextual information, the Differential Selector views the previously selected knowledge $\bm{h}_{k}^{\tau-1}$ as query, and the knowledge sentence $\bm{r}^{\tau}_i$ with its differential information $\bm{o}^{\tau}_i$ as key, to estimate a difference-aware selection distribution:
\begin{align}
  \beta^{\tau}_{\mathrm{Diff}, i} = \bm{v}^\mathrm{T}\tanh\left(\bm{W}_{\mathrm{que}} \bm{h}_{k}^{\tau-1} + \bm{W}_{\mathrm{key}} \left[ \bm{r}_i^{\tau} ; \bm{o}^{\tau}_i \right] \right),
\end{align}
where $\bm{v}$, $\bm{W}_{\mathrm{que}}$ and $\bm{W}_{\mathrm{key}}$ are trainable parameters. 

The final selection distribution is the summation of the distributions of two selectors:
\begin{align}
  \beta^{\tau}_i = \beta^{\tau}_{\mathrm{Ctx}, i} + \beta^{\tau}_{\mathrm{Diff}, i}. \label{equ:addition}
\end{align}
Note that the Differential Selector relies on the previously selected knowledge, thus at the first turn, we set $\beta^{\tau}_{\mathrm{Diff}, i}$ to 0 for each $i$. 




\subsubsection{Selecting Knowledge}

Finally, either adopting the Fused or Disentangled Selection module, the model selects the knowledge sentence with the highest attention score, and uses its representation for further generation\footnote{The model is trained with teacher forcing, where the golden selected knowledge $\bm{h}^{\tau}_{k,{i^{\tau}}^*}$ is used during training.}:
\begin{align}
  \alpha^{\tau}_i & = \mathrm{softmax}_i \left( \beta^{\tau}_i \right), \\
  \widehat{i}^{\tau} = & \mathop{\arg\max}_i \alpha^{\tau}_i ,\ \bm{h}_{k}^{\tau} \triangleq \bm{h}^{\tau}_{k,\widehat{i}^{\tau}}.
\end{align}

\subsection{Decoder}
\label{sec:decoder}

The decoding state is updated by a GRU:
\begin{align}
    \bm{s}_t = \bm{GRU}_D &\left( \bm{s}_{t-1}, \left[ \bm{e}\left(\mathrm{y}_{t-1}\right); \bm{h}_{k} \right] \right), \\
    \bm{s}_0 = \bm{W}_D & \left[ \bm{h}_{c}; \bm{h}_{k} \right]+\bm{b}_D,
\end{align}
where $\bm{W}_D$ and $\bm{b}_D$ are trainable parameters, and $\bm{e}\left(\mathrm{y}_{t-1}\right)$ denotes the embedding of the word $\mathrm{y}_{t-1}$ generated in the last time step.
Then, the decoder outputs the generation probability over the vocabulary (without normalization):
\begin{align}
  {\phi}^G({y}_t=\mathrm{w}) = \mathbf{w}^\mathrm{T}\left(\bm{W}_G\bm{s}_t + \bm{b}_G\right),
\end{align}
where $\bm{W}_G$ and $\bm{b}_G$ are trainable parameters, and $\mathbf{w}$ is the one-hot vector of the word $\mathrm{w}$.
Meanwhile, a copy mechanism \cite{gu-etal-2016-incorporating} is adopted to output additional copy probability of the words in the selected knowledge sentence $\mathrm{k}_{\widehat{i}}$ (without normalization):
\begin{align}
  \phi^C \left( {y}_t = \mathrm{w} \right) =  \sum_{ j: \mathrm{k}_{\widehat{i},j} = \mathrm{w} } \left( \bm{s}_t \right)^\mathrm{T} \bm{H} \left( \bm{h}_{k,\widehat{i}, j} \right),
\end{align}
where $\bm{H}$ is a fully connected layer activated with $\mathrm{tanh}$. 
The final probability distribution is computed as follows:
\begin{align}
  \mathcal{P}\left( {y}_t\! =\! \mathrm{w} \right)\! =\! \frac{1}{Z} \left( \mathrm{e}^{ \phi^G \left( {y}_t = \mathrm{w} \right) }\! +\! \mathrm{e}^{ \phi^C \left( {y}_t = \mathrm{w} \right) } \right)\!,
\end{align}
where $Z$ is the normalization term. 
Then we select the word from vocabulary with the highest probability, saying:
    $\mathrm{y}_t = \mathop{\arg\max}_{\mathrm{w}} \mathcal{P}({y}_t = \mathrm{w})$.

\subsection{Loss}

The negative log likelihood loss is adopted:
\begin{align}
    \mathcal{L}_\mathrm{NLL} = - \sum_{\tau=1}^{T} \sum_{t=1}^{ \left| \mathrm{y}^{\tau} \right|} \log \mathcal{P} \left( {\mathrm{y}^{\tau}_{t}}^* \right),
\end{align}
where ${\mathrm{y}^{\tau}_{t}}^*$ denotes the $t$-th word in the golden response at the $\tau$-th turn and $T$ is the length of turns in the whole dialogue.
We also add supervision on the final knowledge selection distribution:
\begin{align}
  \mathcal{L}_\mathrm{KS} = - \sum_{\tau=1}^{T} \log \alpha^{\tau}_{{i^{\tau}}^*},
\end{align}
where ${i^{\tau}}^*$ denotes the index of the golden selected knowledge sentence at the $\tau$-th turn.
The total loss is their summation:
\begin{align}
    \mathcal{L} = \mathcal{L}_\mathrm{NLL} + \lambda \mathcal{L}_\mathrm{KS}.
\end{align}
where we set $\lambda=1$ in our experiments.

\section{Experiments}

\subsection{Datasets}
\label{sec:datasets}

We evaluated our method on two widely used benchmarks:
 Wizard of Wikipedia (WoW)  \cite{dinan2018wizard}, and  Holl-E \cite{moghe-etal-2018-towards}.

WoW \cite{dinan2018wizard} 
contains multi-turn knowledge-grounded conversations, collected by wizard-apprentice mode. 
Each utterance of the wizard is grounded to a selected knowledge sentence, or indicated by that no knowledge is used.
The dialogues are split into 18,430/1,948/965/968 for Train/Dev/Test Seen/Test Unseen respectively, with 4 turns per dialogue and 61 provided knowledge sentences per turn on average.
Note that the test data is split into \textit{Test Seen} (in-domain) and \textit{Test Unseen} (out-of-domain), where Test Unseen contains topics that are never seen in Train or Dev.

Holl-E \cite{moghe-etal-2018-towards} 
contains conversations in which one speaker is strictly instructed to give utterances by copying or modifying sentences from the given background document.
Similarly, each utterance is annotated regarding the selected knowledge.
Following \citet{Kim2020Sequential}, we tokenized the background document into sentences, and meanwhile ensured that the annotated span is included in a whole sentence.
The dialogues are split into 7,211/930/913 for Train/Dev/Test respectively, with 5 turns per dialogue and 60 provided knowledge sentences per turn on average.


\subsection{Baselines}

We compared our models with the following typical knowledge selection baselines:


\noindent \textbf{MemNet} \cite{ghazvininejad2018knowledge} stores knowledge sentences in its memory units, which are attentively read during decoding. 
We also evaluated a variant (\textbf{MemNet+}) where knowledge selection is supervised by the same $\mathcal{L}_{\mathrm{KS}}$ as our models do.

\noindent \textbf{PostKS} \cite{postks-ijcai} estimates two knowledge selection distributions, where the prior distribution is based on only the context and the posterior one on both the context and the golden response, and their KL divergence is minimized during training.
The knowledge selection of PostKS is supervised by a BOW loss.
We also evaluated two variants, where one uses ${\mathcal{L}_{\mathrm{KS}}}$ instead of the BOW loss to supervise knowledge selection (\textbf{PostKS+}), and the other is further equipped with copy mechanism (\textbf{PostKS++}).

\noindent \textbf{SLKS} \cite{Kim2020Sequential} improves PostKS by using two separate GRUs to update the states of dialog history and previously selected knowledge sentences respectively.
For fair comparison, we replaced the pretrained BERT \cite{devlin-etal-2019-bert} encoder and the Transformer \cite{transformer} decoder in SLKS with BiGRU and GRU respectively, and adopted the same copy mechanism in SLKS as in our models.


\begin{table}[t]
  \centering
  \scalebox{0.8}{
    \begin{tabular}{l|cccc}
      \toprule
      \textbf{Models} & \textbf{ACC} & \multicolumn{2}{c}{\textbf{BLEU-2/4}} & \textbf{ROUGE-2} \\
      \midrule
      \multicolumn{5}{c}{\textbf{WoW Seen}} \\
      \midrule
      \textbf{MemNet} & 13.2**  & 6.6**  & 1.8**  & 3.2**   \\
      {\ \ \ \textbf{+}$\bm{\mathcal{L}_\mathrm{KS}}$} & 18.4**  & 7.2**  & 1.9**  & 3.3**   \\
      \textbf{PostKS} & 13.8**  & 6.9**  & 1.8**  & 3.2**  \\
      {\ \ \ \textbf{+}$\bm{\mathcal{L}_\mathrm{KS}}$} & 22.5**  & 7.5**  & 2.3**  & 3.7**  \\
      {\ \ \ \ \ \ \textbf{+Copy}} & 21.9**  & 9.9**  & 4.5**  & 5.6**   \\
      \textbf{SLKS} & 23.4**  & 11.3  & 5.5  & 6.5   \\
      \cmidrule{1-5}
      \textbf{DiffKS$_\text{Fus}$} & \textbf{25.5}  & \textbf{11.6}  & \textbf{5.7}  & \textbf{6.8} \\
      \textbf{DiffKS$_\text{Dis}$} & 24.7  & 11.3  & \textbf{5.7}  & \textbf{6.8}    \\
      \midrule
      \multicolumn{5}{c}{\textbf{WoW Uneen}} \\
      \midrule
      \textbf{MemNet}  & 12.8**  & 5.7**  & 1.2**  & 2.3**   \\
      {\ \ \ \textbf{+}$\bm{\mathcal{L}_\mathrm{KS}}$}  & 15.9**  & 5.9**  & 1.3**  & 2.3**   \\
      \textbf{PostKS}  & 13.6**  & 5.5**  & 1.2**  & 2.1**   \\
      {\ \ \ \textbf{+}$\bm{\mathcal{L}_\mathrm{KS}}$} & 15.8**  & 6.6**  & 1.5**  & 2.6**  \\
      {\ \ \ \ \ \ \textbf{+Copy}} & 14.9**  & 7.9**  & 3.2**  & 3.9**  \\
      \textbf{SLKS} &  14.7**  & 8.7**  & 3.7**  & 4.6**   \\
      \cmidrule{1-5}
      \textbf{DiffKS$_\text{Fus}$} & \textbf{19.7}  & \textbf{10.0}  & \textbf{4.7}  & \textbf{5.6}   \\
      \textbf{DiffKS$_\text{Dis}$}  & 18.3*  & 9.6  & 4.5  & 5.3   \\
      \midrule
      \multicolumn{5}{c}{\textbf{Holl-E}} \\
      \midrule
      \textbf{MemNet} &  5.1**  & 8.0**  & 4.5**  & 8.9**   \\
      {\ \ \ \textbf{+}$\bm{\mathcal{L}_\mathrm{KS}}$} & 25.1**  & 7.7**  & 4.3**  & 9.0**   \\
      \textbf{PostKS} & 6.1**  & 6.9**  & 3.9**  & 8.6**   \\
      {\ \ \ \textbf{+}$\bm{\mathcal{L}_\mathrm{KS}}$} & 29.5**  & 15.9**  & 8.2**  & 13.1**   \\
      {\ \ \ \ \ \ \textbf{+Copy}} & 28.0**  & 26.5**  & 22.4**  & 23.1**   \\
      \textbf{SLKS}  & 28.6**  & 28.5**  & 24.5**  & 24.9*   \\
      \cmidrule{1-5}
      \textbf{DiffKS$_\text{Fus}$} & 33.0  & 29.5  & 25.5  & 25.9   \\
      \textbf{DiffKS$_\text{Dis}$} &  \textbf{33.5}  & \textbf{29.9}  & \textbf{25.9}  & \textbf{26.4}   \\
      \bottomrule
    \end{tabular}%
  }
  \caption{Automatic evaluation results. 
  The best results are in \textbf{bold}. 
  Significance tests were conducted between the best results and other competitors, with sign test for ACC, bootstrap resampling \cite{koehn-2004-statistical} for BLEU, and Student’s t-test for ROUGE. 
  */** indicate $p$-value $<$ 0.05/0.005 respectively.}
  \label{tab:automatic}%
\end{table}%

\subsection{Implementation Details}

All the models were implemented with PyTorch 
\cite{paszke2017automatic}. 
The sentences were tokenized with NLTK 
\cite{bird-loper-2004-nltk}.
We set the vocabulary size to 20K for WoW and 16K for Holl-E and used the 300-dimensional word embeddings initialized by GloVe \cite{pennington-etal-2014-glove} or from a standard normal distribution $\mathcal{N}(0,1)$. 
We applied a dropout rate of 0.5 on word embeddings. 
The hidden sizes were set to 200 for the encoders (totally 400 for two directions) and to 400 for the decoder.
We adopted the ADAM \cite{kingma2015adam} optimizer with the initial learning rate set to 0.0005.
The batch size was set to 8 dialogues.
{All the models share the same hyperparameter setting} and were trained for 20 epochs on {one NVIDIA Titan Xp GPU}. 
The checkpoints of our reported results were selected according to BLEU-4 on the Dev sets.

\subsection{Automatic Evaluation}

We used several automatic metrics: \textit{\textbf{ACC}}, the accuracy of knowledge selection on the whole test set, corpus-level \textit{\textbf{BLEU-2/4}} \cite{papineni-etal-2002-bleu}, and \textit{\textbf{ROUGE-2}} \cite{lin-2004-rouge}.

As shown in Table \ref{tab:automatic}\footnote{
  We found in \citet{Kim2020Sequential} that BERT usually gives rise to a gain of 2-5 points in ACC, thus our results without using BERT as encoder are within a reasonable range comparing with those in the original reference.
}, our method outperforms significantly all the baselines in all the metrics on three test sets (except BLEU and ROUGE on WoW Seen compared with SLKS), which indicates its superiority in selecting proper knowledge and generating informative responses.
Compared to the baseline models, our models also demonstrate a stronger ability of generalization from in-domain (WoW Seen) to out-of-domain data (WoW Unseen).
It is worth noting that on WoW Unseen, our DiffKS$_\text{Fus}$ obtains a higher accuracy (19.7) of knowledge selection even than the BERT-enhanced SLKS in their original paper (18.3).
We also observed that DiffKS$_\text{Fus}$ performs a bit better on WoW while DiffKS$_\text{Dis}$ on Holl-E.
We conjecture that it is because in Holl-E, the golden selected knowledge among different turns usually has high contextual dependency (for example, they may be continuous sentences in the document), which makes it feasible to predict the next selected knowledge simply conditioned on the differential information.

\begin{table}[t]
  \centering
  \scalebox{0.75}{
    \begin{tabular}{l|ccc|ccc}
      \toprule
      \multicolumn{1}{c|}{} & \multicolumn{3}{c|}{\textbf{Naturalness}} & \multicolumn{3}{c}{\textbf{Appropriateness}}  \\
      \multicolumn{1}{c|}{\textbf{A vs. B}} & \textbf{Win} & \textbf{Lose} & $\bm{\kappa}$ & \textbf{Win} & \textbf{Lose} & $\bm{\kappa}$ \\
      \midrule
      \multicolumn{7}{c}{\textbf{WoW Seen}} \\
      \midrule
      \textbf{Fus / PostKS++} & \textbf{50.3*} & 42.5 & .47  & 4\textbf{9.2*} & 43.1 & .40 \\
      \textbf{Fus / SLKS} & 44.5 & 43.3 & .50      & \textbf{44.0*} & 38.7 & .48 \\
      \textbf{Dis / PostKS++} & \textbf{50.6*} & 44.9 & .42 & \textbf{50.5*} & 44.4 & .38 \\
      \textbf{Dis / SLKS} & 42.7 & 43.8 & .41 & 46.4 & 41.4 & .47 \\
      \textbf{Fus / Dis} & 43.2 & 42.8 & .49 & 39.3 & 40.9 & .57  \\
      \midrule
      \multicolumn{7}{c}{\textbf{WoW Unseen}} \\
      \midrule
      \textbf{Fus / PostKS++} & \textbf{48.8*} & 43.2 & .57 & \textbf{49.3**} & 40.5 & .60  \\
      \textbf{Fus / SLKS} & \textbf{47.9*} & 41.8 & .44 & \textbf{47.3*} & 40.9 & .47 \\
      \textbf{Dis / PostKS++} & \textbf{52.0**} & 36.4 & .46 & \textbf{46.8*} & 39.9 & .49 \\
      \textbf{Dis / SLKS} & \textbf{46.5*} & 39.7 & .45 & \textbf{47.8*} & 42.3 & .47  \\
      \textbf{Fus / Dis} & 39.8 & 42.4 & .52 & 41.5 & 37.8 & .53  \\
      \bottomrule
    \end{tabular}%
  }
  \caption{Human observational evaluation results. Ties are not shown. Significance tests were conducted with sign test. ${\kappa}$ denotes the Fleiss' Kappa which measures annotation agreement.}
  \label{tab:manual}%
\end{table}%

\subsection{Human Observational Evaluation}

We conducted human observational evaluation with pair-wise comparison, where our two models were compared with PostKS++ and SLKS.
100 dialogues were respectively sampled from WoW Seen/Unseen.
For each pair of dialogues generated from two models (suppose with $T$ turns), annotators from Amazon Mechanical Turk were hired to give preferences (win, lose, or tie) for each response pair of all the $T$ turns in terms of different metrics.
Each pair-wise comparison of dialogues was judged by 3 curators. 
We adopted the following two metrics:
\textit{\textbf{Naturalness}} evaluates the fluency and readability of a response. 
\textit{\textbf{Appropriateness}} evaluates the relevance to the context and whether a response contains appropriate knowledge information to the context.

Results are shown in Table \ref{tab:manual}, where the Fleiss' Kappa \cite{fleiss1971measuring} values show almost moderate agreements ($0.4<\kappa<0.6$).
Our models significantly outperform PostKS++ in both metrics, and also generally outperform SLKS in terms of Appropriateness.
Again, the advantage of our models on WoW Unseen is more evident than on WoW Seen.

\begin{table}[t]
  \centering
  \scalebox{0.8}{
    \begin{tabular}{l|cc}
      \toprule
      \textbf{Models} & \textbf{WoW Seen} & \textbf{WoW Uneen} \\
      \midrule
      \textbf{Human}$^{\dagger}$ & 4.13 (1.08) & 4.34 (0.98) \\
      \cmidrule{1-3}
      \textbf{PostKS++} & 2.30 (1.06) & 2.13 (1.10) \\
      \textbf{SLKS} & 2.32 (1.11) & 2.22 (1.15) \\
      \textbf{DiffKS$_\text{Fus}$} & \textbf{2.43} (0.96) & \textbf{2.39} (1.16) \\
      \textbf{DiffKS$_\text{Dis}$} & 2.39 (1.17) & 2.38 (1.19) \\
      \bottomrule
    \end{tabular}%
  }
  \caption{Human interactive evaluation results. 
  The standard deviation is marked in parentheses.
  The results of human$^{\dagger}$ are from \citet{dinan2018wizard,Kim2020Sequential}.
  }
  \label{tab:interactive}%
\end{table}%

\subsection{Human Interactive Evaluation}

We further conducted human interactive evaluation where real humans converse with one model about a specific topic.
We compared PostKS++ and SLKS with our two models.
The workers from Amazon Mechanical Turk were asked to first select one topic from 2-3 provided candidate topics, and then converse with one of the models for 3-5 dialogue turns.
After conversation, they were required to rate the dialog model with a 5-star scale in terms of the fluency and informativeness of the utterances and the coherence of the whole dialog.
Following \citet{dinan2018wizard,Kim2020Sequential}, the interactive evaluation was implemented with ParlAI 
\cite{miller2017parlai}.
For each model, we averaged the scores from 150 collected conversations on each test set of WoW.
We also reported the results of human-human dialog from \citet{dinan2018wizard,Kim2020Sequential}, where each worker converses with another human and the latter has access to knowledge sentences just like the models do.

Results are shown in Table \ref{tab:interactive}\footnote{We found in \citet{dinan2018wizard,Kim2020Sequential} that the stddev values of dialog models are usually between 1.0 and 1.4, thus our results are within a reasonable range.}, where DiffKS$_\text{Fus}$ gains the highest scores and our models both outperform the other two state-of-the-art baselines, indicating that our models are favorably  preferred by human annotators.


\begin{table}[t]
  \centering
  \scalebox{0.8}{
    \begin{tabular}{l|cccc}
      \toprule
      \multicolumn{1}{l|}{\textbf{Models}} & \multicolumn{1}{c}{\textbf{ACC}} & \multicolumn{2}{c}{\textbf{BLEU-2/4}} & \multicolumn{1}{c}{\textbf{ROUGE-2}} \\
      \midrule
      \multicolumn{5}{c}{\textbf{WoW Seen}} \\
      \midrule
      \textbf{DiffKS$_\text{Dis}$} & 24.7  & 11.3  & 5.7  & 6.8   \\
      w/o DiffSel & \underline{22.3**}  & \underline{10.6**}  & \underline{4.9**}  & \underline{5.9**}  \\
      w/o CtxSel & 24.6  & 10.9  & 5.3*  & 6.5  \\
      \midrule
      \multicolumn{5}{c}{\textbf{WoW Unseen}} \\
      \midrule
      \textbf{DiffKS$_\text{Dis}$} & 18.3  & 9.6  & 4.5  & 5.3  \\
      w/o DiffSel & \underline{15.5**}  & \underline{8.8**}  & \underline{3.8**}  & \underline{4.4**}   \\
      w/o CtxSel & 18.4  & 9.1*  & 4.1*  & 5.0   \\
      \midrule
      \multicolumn{5}{c}{\textbf{Holl-E}} \\
      \midrule
      \textbf{DiffKS$_\text{Dis}$} & 33.5  & 29.9  & 25.9  & 26.4   \\
      w/o DiffSel & \underline{29.1**}  & \underline{27.9**}  & \underline{23.8**}  & 25.1   \\
      w/o CtxSel & 31.6**  & 28.4**  & 24.7**  & \underline{24.8*}   \\
      \bottomrule
    \end{tabular}%
  }
  \caption{Ablation tests. The larger performance drops between the two ablation models are underlined. The significance tests are conducted between the ablation models and the complete model DiffKS$_\text{Dis}$. 
  }
  \label{tab:ablation}%
\end{table}

\subsection{Ablation Test}
\label{sec:ablation}

In order to verify the effectiveness of the differential information in knowledge selection, we conducted ablation tests, which were specifically based on the disentangled variant DiffKS$_\text{Dis}$.
In DiffKS$_\text{Dis}$, we removed either the Differential Selector (DiffSel) or the Contextual Selector (CtxSel), and trained the model with only one of the two selectors.

Results are shown in Table \ref{tab:ablation}. 
Without the differential selector, the model performance is remarkably impaired in all the metrics on three test sets, indicating the importance of utilizing differential information.
In comparison, removing the contextual selector is less influential (with less performance drop). 
We conjecture that this may result from the characteristics of datasets. 
For instance, in WoW, the apprentice (without access to knowledge) usually reacts passively to the wizard (having access to knowledge).
Thus the apprentice posts (contextual information) have limited influence in driving the conversation, which is instead affected or controlled by the wizard.
In this case, our differential information that can predict the process of knowledge transition has more influence than the contextual information.
In addition, same as \citet{Kim2020Sequential}, the knowledge sentences in Holl-E are obtained by segmenting a long document into single sentences, which implies that there exists the relevance or contextual dependency between knowledge sentences. 
Consequently, the differential information is still able to provide considerable clues for knowledge selection even without access to the new user post (the context).

\begin{table}[t]
  \centering
  \scalebox{0.8}{
    \begin{tabular}{lc|cccc}
      \toprule
      \textbf{Models} & $\bm{M}$ & \textbf{ACC} & \multicolumn{2}{c}{\textbf{BLEU-2/4}} & \textbf{ROUGE-2} \\
      \midrule
      \multicolumn{6}{c}{\textbf{WoW Seen}} \\
      \midrule
      \multirow{3}{*}{\textbf{DiffKS$_\text{Fus}$}} & \textbf{1} &  25.5 & 11.6  &  5.7  &   6.8      \\
        & \textbf{2} &  \textbf{26.3} & \textbf{11.7}  &  5.8  &   7.0     \\
        & \textbf{3} &  26.1 & 11.6  &  5.7  &   6.9      \\
      \cmidrule{1-6}
      \multirow{3}{*}{\textbf{DiffKS$_\text{Dis}$}} & \textbf{1} &  24.7 & 11.3  &  5.7  &   6.8     \\
        & \textbf{2} &  26.1 & \textbf{11.7}  &  \textbf{6.0}  &   \textbf{7.1}     \\
        & \textbf{3} &  25.0 & 11.1  &  5.7  &   6.7      \\
      \midrule
      \multicolumn{6}{c}{\textbf{WoW Uneen}} \\
      \midrule
      \multirow{3}{*}{\textbf{DiffKS$_\text{Fus}$}} & \textbf{1} &  19.7 & 10.0  &  4.7  &   5.6        \\
        & \textbf{2} &  \textbf{20.4} & \textbf{10.6}  &  \textbf{5.2}  &   \textbf{6.0}      \\
        & \textbf{3} &  19.5 & 9.8   &  4.8  &   5.6      \\
        \cmidrule{1-6}
      \multirow{3}{*}{\textbf{DiffKS$_\text{Dis}$}} & \textbf{1} &  18.3 & 9.6   &  4.5  &   5.3     \\
        & \textbf{2} &  19.4 & 9.9   &  4.6  &   5.3      \\
        & \textbf{3} &  19.1 & 9.9   &  4.5  &   5.2       \\
      \midrule
      \multicolumn{6}{c}{\textbf{Holl-E}} \\
      \midrule
      \multirow{3}{*}{\textbf{DiffKS$_\text{Fus}$}} & \textbf{1} &  33.0 & 29.5  &  25.5  &  25.9        \\
       & \textbf{2} &  33.2 & 30.1  &  26.1  &  26.2      \\
       & \textbf{3} &  33.1 & 30.0  &  26.3  &  26.3      \\
       \cmidrule{1-6}
       \multirow{3}{*}{\textbf{DiffKS$_\text{Dis}$}} & \textbf{1} &  33.5 & 29.9  &  25.9  &  26.4     \\
       & \textbf{2} &  \textbf{33.9} & 31.2  &  \textbf{27.2}  &  \textbf{26.9}     \\
       & \textbf{3} &  33.8 & \textbf{31.3}  &  26.8  &  26.7       \\
      \bottomrule
    \end{tabular}%
  }
  \caption{Comparison between results with different $M$.}
  \label{tab:auto_M}%
\end{table}%

Furthermore, after removing DiffSel, DiffKS$_\text{Dis}$ reduces to a vanilla knowledge selection model where the supervision $\mathcal{L}_\mathrm{KS}$ was directly applied on the `prior' selection distribution.
Nevertheless, the performance of the ablated model is sometimes competitive to the baselines (for instance, in terms of ACC, DiffKS$_\text{Dis}$ w/o DiffSel obtains 22.3/15.5/29.1 vs. 21.9/14.9/28.0 of PostKS++).
It may result from the gap between training and inference caused by the prior-posterior framework adopted in PostKS and SLKS, which may be not superior over directly training the prior selection distribution\footnote{
  The prior-posterior framework is first proposed by PostKS without direct supervision $\mathcal{L}_\mathrm{KS}$ on knowledge selection. 
  While in this paper and \citet{Kim2020Sequential} the supervision $\mathcal{L}_{\mathrm{KS}}$ is available, the prior-posterior framework may not be superior any more.
}.

\section{Discussion}

\subsection{Difference From More Turns}

To investigate the impact of increasing the turns of differential information (the $M$ in Equ.\ref{equ:diff_M}), we additionally experimented with $M=2,3$, and took the arithmetic average for simplicity in Equ.\ref{equ:diff_M}, saying $\forall i, \lambda_i=1/M$.

Results are shown in Table \ref{tab:auto_M}. 
We can find that $M=2$ generally achieves the best performance compared with $M=1,3$ for both DiffKS$_\text{Fus}$ and DiffKS$_\text{Dis}$ (while $M=3$ is still better than $M=1$). 
It further turns out the effectiveness of explicitly modeling differential information.
We also conjecture that the model performance would be further improved by assigning the nearest/farthest difference with the largest/smallest weight in Equ.\ref{equ:diff_M}, saying $\lambda_1 > \lambda_2 > \dots > \lambda_M$, which is more reasonable than the simplified arithmetic average.

\begin{table}[t]
  \centering
  \scalebox{0.8}{
    \begin{tabular}{l|ccccc}
      \toprule
      \textbf{Models} & \textbf{1$^\text{st}$} & \textbf{2$^\text{nd}$} & \textbf{3$^\text{rd}$} & \textbf{4$^\text{th}$} & \textbf{5$^\text{th}$} \\
      \midrule
      \multicolumn{6}{c}{\textbf{WoW Seen}} \\
      \midrule
      \textbf{PostKS++} & 56.8 & 15.6 & 9.6 & 6.2 & 4.1  \\
      \textbf{SLKS} & \textbf{57.4} & 18.4 & 10.1 & 8.9 & 5.4  \\
      \textbf{DiffKS$_\text{Fus}$} & \textbf{57.4} & \textbf{22.5} & \textbf{12.8} & 9.8 & 7.4 \\
      \textbf{DiffKS$_\text{Dis}$} & 56.6 & 21.5 & 11.2 & \textbf{10.2} & \textbf{7.9}  \\
      \midrule
      \multicolumn{6}{c}{\textbf{WoW Uneen}} \\
      \midrule
      \textbf{PostKS++}  & 42.8 & 8.5 & 4.1 & 4.8 & 4.6   \\
      \textbf{SLKS} &  \textbf{43.0} & 6.1 & 5.2 & 4.9 & 5.0 \\
      \textbf{DiffKS$_\text{Fus}$} & 40.9 & \textbf{21.2} & \textbf{10.5} & \textbf{7.7} & 4.6  \\
      \textbf{DiffKS$_\text{Dis}$}  & 40.2 & 16.1 & 10.3 & \textbf{7.7} & \textbf{6.1}  \\
      \midrule
      \multicolumn{6}{c}{\textbf{Holl-E}} \\
      \midrule
      \textbf{PostKS++} & 62.8 & 17.9 & 18.8 & 20.0 & 23.2  \\
      \textbf{SLKS}  & 65.2 & 18.4 & 19.2 & 21.3 & 19.6  \\
      \textbf{DiffKS$_\text{Fus}$} & \textbf{65.8} & 22.3 & 22.1 & 25.5 & 25.8 \\
      \textbf{DiffKS$_\text{Dis}$} & 63.9 & \textbf{23.0} & \textbf{23.4} & \textbf{26.0} & \textbf{28.3}   \\
      \bottomrule
    \end{tabular}%
  }
  \caption{Knowledge selection accuracy over turns.}
  \label{tab:acc_over_turn}%
\end{table}%

\subsection{Accuracy Over Turns}
\label{sec:acc_over_turn}

To verify whether the sequential knowledge selection facilitates knowledge selection in later turns,
we evaluated the accuracy of knowledge selection at different turns.
The statistics are shown in Table \ref{tab:acc_over_turn}.
Our two models have the highest accuracy from the 2$^\text{nd}$ to 5$^\text{th}$ turns and outperform SLKS and PostKS++ (and SLKS also generally outperforms PostKS++). The results show that our models can select more accurate knowledge consistently over different turns.

\subsection{Case Study}

\begin{figure}[ht]
  \centering
  \includegraphics[width=\linewidth]{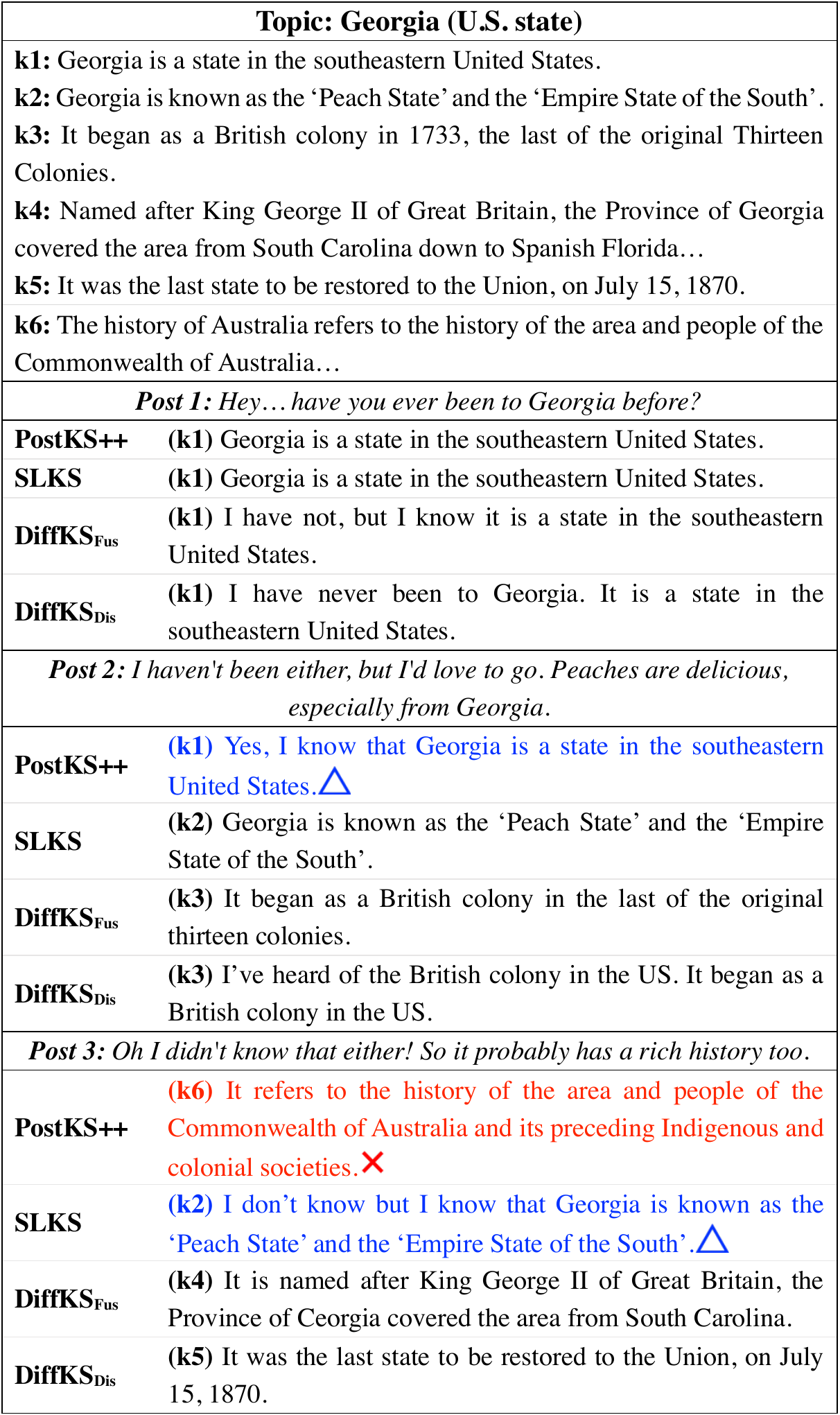}
  \caption{Case study. We marked the selected knowledge sentence in parentheses before each response.
  The knowledge k1-k5 are about the topic \textit{Georgia (U.S. state)}, while k6 is about \textit{History of Australia}.
  The {\color{blue} \textbf{blue} $\bm{\triangle}$} denotes duplicate responses resulting from repetitive knowledge selection.
  The {\color{red} \textbf{red} $\bm{\times}$} denotes incoherent responses resulting from selecting a far different knowledge from previous turns.
  } 
  \label{fig:case_study}
\end{figure}

We show a case from WoW Seen in Figure \ref{fig:case_study}, which compares the responses generated by PostKS++, SLKS and our two models.

At the 2$^\text{nd}$ turn, PostKS++ generates almost the same responses as at the 1$^\text{st}$ turn due to the repetitive knowledge selection. 
Similar cases occur for SLKS at the 2$^\text{nd}$ and the 3$^\text{rd}$ turns.
Moreover, PostKS++ selects a quite different knowledge sentence at the 3$^\text{rd}$ turn from those at previous turns, which is about the topic \textit{History of Australia} but not \textit{Georgia (U.S. state)}.
As a result, PostKS++ generates a response which is not coherent to the previous context at the 3$^\text{rd}$ turn.
In contrast, our two models select both diverse and appropriate knowledge sentences at all the turns, thereby generating informative responses and making the dialog coherent and natural.

\section{Conclusion}

We present a novel difference-aware knowledge selection method for multi-turn knowledge-grounded conversation generation.
Our method first compares the candidate knowledge provided at the current turn with the previously selected knowledge, and then selects appropriate knowledge to be used in generation. 
Experimental results show that our method is able to select knowledge more accurately and to generate more informative responses, outperforming significantly the state-of-the-art baselines.

\section*{Acknowledgments}
This work was jointly supported by the NSFC projects (Key project with No. 61936010 and regular project with No. 61876096), and the Guoqiang Institute of Tsinghua University, with Grant No. 2019GQG1.
We thank THUNUS NExT Joint-Lab for the support.

\bibliography{anthology,emnlp2020}
\bibliographystyle{acl_natbib}

\end{document}